\documentclass[letterpaper, 10 pt, conference]{ieeeconf}  

\IEEEoverridecommandlockouts                              

\makeatletter
\let\NAT@parse\undefined
\makeatother

\usepackage{graphics} 
\usepackage{epsfig} 
\usepackage{mathptmx} 
\usepackage{times} 
\usepackage{amsmath} 
\usepackage{amssymb}  
\usepackage{xcolor}
\usepackage{url}
\usepackage{lineno}
\usepackage{setspace}
\usepackage{cite} 
\usepackage{hyperref}
\usepackage{subcaption}

\usepackage{algorithmic}
\usepackage{algorithm}
\makeatletter
\newenvironment{pseudocode}[1][htb]{%
    \renewcommand{\ALG@name}{Algorithm 1. Sensitivity-Guided Stabilization Control}
   \begin{algorithm}[#1]%
  }{\end{algorithm}}

\newenvironment{pseudocodeLE}[1][htb]{%
    \renewcommand{\ALG@name}{Algorithm 2. Estimate Truncated Lyapunov Exponents}
   \begin{algorithm}[#1]%
  }{\end{algorithm}}  
  
\makeatother

\title{\LARGE \bf
SuPLE: Robot Learning with Lyapunov Rewards  
}

\author{Phu Nguyen$^{1}$, Daniel Polani$^{2}$, Stas Tiomkin$^{3}$
\thanks{$^{1}$Computer Engineering Dept., Davidson College of Engineering, San Jose State University, CA 95192, USA;
        {\tt\small phu.c.nguyen@sjsu.edu}}%
\thanks{$^{2}$  Department of Computer Science, University of Hertfordshire, AL10 9AB, UK;
        {\tt\small d.polani@herts.ac.uk}}%
\thanks{$^{3}$ Department of Computer Science, Texas Tech University, TX 79409, USA;
        {\tt\small stas.tiomkin@ttu.edu}; corresponding author}%
}

\begin{document}

\maketitle
\thispagestyle{empty}
\pagestyle{empty}

\begin{abstract}
The reward function is an essential component in robot learning. Reward directly affects the sample and computational complexity of learning, and the quality of a solution. The design of informative rewards requires domain knowledge, which is not always available. We use  the properties of the dynamics to produce system-appropriate reward without adding external assumptions. Specifically,  we explore an approach to utilize the Lyapunov exponents of the system dynamics to generate a system-immanent reward. We demonstrate that the \textit{Sum of the Positive Lyapunov Exponents} (SuPLE) is a strong candidate for the design of such a  reward. We develop a computational framework for the derivation of this reward, and demonstrate its effectiveness on  classical benchmarks for sample-based stabilization of various dynamical systems. 
It eliminates the need to start the training trajectories at arbitrary states, also known as auxiliary exploration. 
While the latter is a common practice in simulated robot learning,  it is unpractical to consider to use it in real robotic systems, since they typically start from natural rest states such as 
a pendulum at the bottom, a robot on the ground, etc.
and can not be easily initialized at arbitrary states. Comparing the performance of SuPLE to  commonly-used reward functions, we observe that the latter fail to find a solution without auxiliary exploration, even for the task of swinging up the double pendulum and keeping it stable at the upright position, a prototypical scenario for multi-linked robots. SuPLE-induced rewards for robot learning offer a novel route for effective robot learning in  typical as opposed to highly specialized or fine-tuned scenarios.
Our code is publicly available for reproducibility and further research.    
\end{abstract}

{
\section{Introduction}
The stabilization of dynamical systems is a central task in many
applications, including robotic control. The design of appropriate
control rules can require quite substantial domain knowledge \cite{1383932,ASTROM2000287,BORTOFF19962810}. It is
indeed one of the major contributions of modern AI methodologies, such
as Reinforcement Learning (RL), to reduce the required explicit injection
of such knowledge \cite{sutton2018reinforcement,8623253,10156044}. However, even there, the performance of an
effective learner is usually boosted by a well-designed reward model.
For this purpose, one uses reward shaping  \cite{ng_shaping,8623253,10156044,10587783}, but
that itself requires domain knowledge. Where this is not the case, the
learner runs the risk to be inefficient. This is due to the slow process of
needing the minimally specified reward structure to be propagated
throughout the configuration space through the learning algorithm. A frequently used remedy for ineffective reward in robot learning  is to randomly reset the agent to arbitrary states in state space. This  boosts the exploration of the state space artificially, and allows one to train even with sparse rewards, where the agent receives a positive reward on task completion only (and zero reward elsewhere). 

Such \emph{auxiliary exploration} is possible in simulation, but it is impractical for training real robotic systems, because it requires the, say,  multi-linked robotic system to be regularly physically reset to random states to learn the task of self-righting and stabilization at the upright position, one of the most important tasks for robot learning. 

Here, we propose an alternative route to achieve stabilization which
requires only minimal external domain knowledge beyond the dynamical
system itself. Concretely, we use truncated Lyapunov exponents to
reward the controller, who learns to direct the system towards
maximally sensitive states. This turns out to be surprisingly
effective in identifying states of interest and providing a natural
reward function which, except for the specification of the free parameter determining the depth of the time
horizon, only requires the system dynamics itself. In particular,
it does not need a human to introduce detailed prior understanding
about a reward function. 

Amongst the models of interest are classic benchmarks such as 
variations of differing complexity of the pendulum swing-up task. While, for
the single pendulum, many working solutions are known, the design of
stabilizing controllers for more complex scenarios, such as the double
pendulum, is more intricate and any learning of the controller is
substantially more difficult; highly elaborate and optimized methods
such as PILCO \cite{deisenroth2011pilco} can be used for this.

Ideally, to be effective, a reward structure should be both \emph{dense} (it is better to receive informative
rewards throughout the run rather than, say, only at the target
region) and
\emph{informative} (it should ideally help to directly specify what actions
to select rather than extracting that information piecemeal). A well
set-up RL system will, after convergence, ultimately build a suitable value or $Q$-function
from the rewards, but if the rewards themselves can be
imbued with domain knowledge, this process can be sped up drastically.

However, especially in larger search spaces, it is not obvious how to
set up an appropriate reward. Easy-to-specify  rewards in such
systems will typically be quite sparse, e.g.\ by mostly distinguishing
target from non-target regions. Other rewards might be based on naive reduction of distance to the target region. 
Thus, for effective control, this reward will first have to
be 
propagated back throughout the whole system with a suitable RL algorithm to become
informative and useful for the controller. An effective  reward function would thus ideally already encapsulate important information about the structure of the control task.

In the present paper, we will show that the truncated Lyapunov
exponent is a prime candidate for such a reward function. Utilizing it
as reward signal, we will show that a resulting controller is able,
for instance, to execute both swing-up as well as the stabilization in
a pendulum scenario. While this Lyapunov exponent can be typically
computed from models, in principle, it can also be estimated from samples \cite{wolf1985determining}.
Trained with the truncated Lyapunov exponents as reward, our controller
guides the system to its most sensitive state, the upright positions. 
Crucially, we observe that above exponents induce larger basins of
attraction around the topmost sensitive states, thus the
controller can follow the trajectory to these states even from far away.

We use  
truncated (i.e., local) Lyapunov exponent spectra as surrogate reward
signal replacing any externally specified one. This reward signal then drives an efficient and
scalable algorithm to carry out the ``natural'' control task of stabilization. We demonstrate the advantages of the
proposed method over the standard data-driven control methods in a
number of scenarios. Specifically, the results will show that our data-driven
method is capable to efficiently construct optimal controllers without
manual engineering of the payoff/reward function. Finally, we will discuss reasons why  this method works, and its
erstwhile limitations at the present stage.

All the experiments and results can be reproduced by our code repository, which we make publicly available at \url{https://github.com/phunguyen1195/TruncLE}

\section{Prior Work}

Information-theoretic methods to generate intrinsic motivations to control various scenarios have been used for a number of years   \cite{dai2023empowerment,qureshi2018adversarial,kwon2021variational,zhao2020efficient,zhao2019avoiding,salge2017empowerment,mohamed2015variational,sharma2019dynamics,eysenbach2018diversity,salge2014empowerment,klyubin2005empowerment,jung2011empowerment,klyubin08:_keep_your_option_open}. One insight is that the maximal potential mutual information between actions and observations acts as a meaningful signal for learning nontrivial behaviour. One  disadvantage is the lack of their interpretability in intrinsic terms  of the system's dynamics, such as the Lyapunov exponents. 
However, such links clearly exist, for instancehe following  fundamental result connecting mutual information in Markov processes with discrete state space and their Lyapunov exponents \cite{holliday2002entropy,holliday2005shannon}. 

More recently, a connection was established between  Lyapunov exponents and information-theoretic methods for intrinsic motivation \cite{tiomkin2024intrinsic} such as empowerment \cite{klyubin2005empowerment,salge2017empowerment,zhao2020efficient}, causal entropic forcing \cite{wissner2013causal}, variational intrinsic control \cite{gregor2016variational}, and other. In this work, we exploit this connection for the generation of informative reward in robot learning.

Specifically, we focus on finding and stabilizing the unstable equilibria in dynamics, crucial for up-right locomotion, self-righting and related tasks.

\section{Preliminaries}
In this section, we review the necessary background for our method, consisting of the truncated Lyapunov spectra and RL.

\subsection{Lyapunov Spectra of Dynamical Systems}

To estimate Lyapunov Exponents, (LE), for a known dynamics, $f$, with $n$-dimensional state $s\in {S}$: 
\begin{align}
s_{t+1}=f(s_t),
\end{align}
we consider the evolution of the principal axes, ($p_1(0), p_2(0), \dots, p_n(0)$), of the $n$-dimensional sphere at $t=0$ in the dynamics state space, $S$. The $i$-th LE, $\lambda_i$, is defined by \cite{sandri1996numerical,wolf1985determining,strogatz2018nonlinear,shaw1981strange} 
\begin{align}
\forall i\;:\;\lambda_i& = \underset{t\rightarrow \infty}{\mbox{ lim }} \frac{1}{t}\log_2 \biggl(\frac{p_i(t)}{p_i(0)}\biggr),\label{eq:LE}&&\text{(the $i$-th exponent)}\\
\Lambda& = (\lambda_1, \lambda_2, \dots, \lambda_n)&&\text{(Lyapunov Spectra)}\\
&\mbox{where }\lambda_i \ge \lambda_{i+1}\nonumber
\end{align}
and $p_i(t)$ is the norm of the $i$-th axis at time $t$. We will denote the $i$-th positive LE by $\lambda_i^+$, thus the maximal LE is $\lambda_1^+$, and the sum of positive LE is:
\begin{align}
\mbox{SuPLE} \doteq \sum_{i}\lambda_i^+. \label{eq:SuPLE}   
\end{align}

The positive LEs are related to
the expansion of the state space volume, while the negative ones are related to its contraction \cite{wolf1985determining}. Thus, the sum of positive LE, SuPLE, reflects the total volume expansion. In practice, the direction, $p_1$, associated with the largest positive LE, $\lambda_1$, often dominates the other directions $p_i$, which collapse on $p_1$, hampering the estimation of their corresponding exponents, $\lambda_i$. The remedy to this collapse is to orthogonalize the set ($p_1(0), p_2(0), \dots, p_n(0)$) by the Gram-Schmidt orthogonalization process at a pre-defined number of time steps. 

The Lyapunov spectra, $\Lambda$, defined in Eq.~\eqref{eq:LE} do not depend on the initial state for $t\rightarrow \infty$. State dependent spectra, $\Lambda(s)$, however, are calculated with a finite time, and are denoted by the truncated LE \cite{TRUNCLYAP}.
Both $\Lambda$ and $\Lambda(s)$ can be efficiently estimated from data \cite{wolf1985determining}, which we utilize in this work. 
 To formulate {\it informative dense reward} for sensitivity-guided
stabilization, we hypothesize that {\it truncated-in-time Lyapunov
  exponents} provide useful state-dependent information for the derivation of stabilizing controllers by data-driven methods. The truncated LE is calculated for a finite number of steps, which is in contrast to the global LE derived for $t'\rightarrow \infty$.

To evaluate the effectiveness of SuPLE, we couple it with SOTA methods in RL, whose basics we briefly overview in the next section.

{
\subsection{Reinforcement Learning (RL)}
The RL setting is modeled as a Markov
Decision Process (MDP) defined by: the state space $S$, the action
space $A$, the transition probabilities $p(s'\mid s, a)$, the initial
state distribution $p_0(s)$, the reward function $r(s,a) \in R$, which
is typically manually designed for a particular task, and the discount
factor $\gamma$. The goal of RL is to find an optimal control policy
$\pi(a\mid s)$ that maximizes the expected return, defined by \cite{sutton2018reinforcement, haarnoja2018soft}
\begin{align}
\underset{\pi}{\mbox{ max }}\mathrm{E}_{s_0\sim p_0,
  a\sim\pi, s'\sim p}\sum_{t=1}\gamma^{t-1} r(s_t, a_t).
\end{align}

\subsection{Soft Actor-Critic}

In more complicated environments, especially, where enhanced exploration is required, the {\it Soft-Actor Critic} \cite{haarnoja2018soft} (SAC) is a more appropriate method. The SAC algorithms augment reward with the entropy of policy, which explicitly encourages exploration:
\begin{align}
\underset{\pi\in \mathcal{P}}{\mbox{ max }}\mathrm{E}_{s_0\sim p_0, a\sim\pi, s'\sim p}\sum_{t=1}\gamma^{t-1} \bigl(r(s_t, a_t) - \log \pi(a_t\mid s_t)\bigr)\label{eq:SAC},
\end{align}

The essential component in any RL objective is reward, which is
usually designed manually, rather than based on the important
properties of dynamics, as in the present work. Here, we show that the
truncated LEs provide the agent with {\it informative
and dense reward} without reward engineering and/or shaping.

The commonly used rewards functions are `i', the sparse reward, which provides the learner with reward `1' in the target state, and with reward `0' elsewhere, and `ii', the error between the current state and the target state, measured by a particular norm.  

Usually, the training involves auxiliary exploration, where a simulated robot is repeatedly initialized in arbitrary states, which is impractical when training  real robotic systems. Auxiliary exploration is crucial for the commonly-used rewards even in such powerful algorithms in RL as SAC and PPO \cite{schulman2017proximal}. To demonstrate the strengths of the proposed reward, SuPLE, we plug it into the RL algorithms in our experiments, and show that SuPLE is informative enough to solve complicated tasks without auxiliary exploration, while the commonly used above-mentioned rewards (`i' and `ii') can not solve these tasks.

}

\section{Proposed Method}
Given a nonlinear dynamics $f(s)$ and its Jacobian $J(s)$ we estimate the truncated LE at the state $s$ by propagating an orthonormal frame $V=\{ v_1, v_2, ..., v_n \}$ for $T$ time steps \cite{wolf1985determining} as summarized in Algorithm `1' and explained below.

As mentioned in Section II A, each vector in the evolved frame, $V$, tends to fall along the local direction of most rapid growth, which corresponds to $v_1$. To prevent this collapse to the dominant direction, we re-orthogonalize the frame, $V$, by the Gram-Schmidt orthogonalization process. {This way $\log_2 \bigl(||v_i||_2\bigr)$ represents the change of the state space volume in the $i$-th direction in a single propagation step by $J(s)$ with (small) stepsize $dt$. These changes are accumulated over time to yield the total change of the $i$-th axis over the time interval $t=Tdt$, denoted by $p_i(t)$ in Eq.~\eqref{eq:LE}.}

Here, Gram-Schmidt re-orthogonalization process replaces the evolved frame $V$ to a corresponding orthogonal set of vectors. Grand-Schmidt process re-orthogonalizes the axes while keeping the direction of the fastest growing vector in the system. The accumulated $\lambda_i$ is divided by $t=Tdt$.

 In  \emph{Algorithm 1} below, we combine truncated LE with the formalism of RL.
\noindent The data on line 3 in \emph{Algorithm 1} is collected with the current policy, $\pi$. In line 4, $\Lambda^{\pi}$ is estimated by \emph{Algorithm 2} at the states visited by $\pi$. That can be done either by model-based LE estimation as in the current work, or by the existing efficient sample-based methods \cite{wolf1985determining,dieci1997compuation,sandri1996numerical}. The notation on line 6 of \emph{Algorithm 1} is the policy improvement induced by the SuPLE reward, using Soft-Actor Critic, Eq.~\eqref{eq:SAC}.

		\begin{pseudocode}[h!] 
		\renewcommand\thealgorithm{} \caption{} 
			\begin{algorithmic}[1]
   \setstretch{1.35}
				\STATE initialize the policy, $\pi:{S}\rightarrow {A}$
				\REPEAT
				\STATE $\tau^{\pi}\doteq\{{s}_t, {a}_t\}_{t=1}^T\leftarrow\mbox{ collect data with }\pi$
				\STATE $\bigl\{{\Lambda}^{\pi}\bigr\}_{t=1}^T\leftarrow \mbox{{EstimateTruncatedLE}}(\tau^{\pi})$
    \STATE $\bigl\{\mbox{SuPLE}\bigr\}_{t=1}^T\gets\bigl\{{\Lambda}^{\pi}\bigr\}_{t=1}^T$\hfill\COMMENT{Eq.~\eqref{eq:SuPLE}}
				\STATE $\pi\leftarrow\mbox{ improve $\pi$ with }\{{s}_t, {a}_t, \mbox{SuPLE}_t\}_{t=1}^T \mbox{ by SAC}$
				\UNTIL{convergence}
				\RETURN{\small$\pi^*$, stabilizing at the states with the maximal LE}
			\end{algorithmic} 
   \label{alg:stabilization_control}
		\end{pseudocode}

Via this scheme, we use the LEs of dynamical systems to determine rewards instead of relying on human knowledge or reward engineering.  

\begin{pseudocodeLE}[h] 
		\renewcommand\thealgorithm{} \caption{} \label{algo:TRLE}
\begin{algorithmic}[1]
\setstretch{1.35}
\REQUIRE $s$-state, $f(s)$-dynamics, $J(s)$-Jacobian, $T$-horizon.
\STATE \textbf{init}: 
\STATE\qquad$V=(v_1, v_2,\dots, v_n)$\hfill\COMMENT{an orthonormal frame at $s$}
\STATE\qquad$\forall i\;:\; \lambda_i= 0$
\FOR {$t\in[1, \dots, T]$} 
  \STATE $\forall i\;:\;v_i=J(s)v_i$\hfill\COMMENT{propagate $V$ vector by vector}
  \STATE $s=f(s)$\hfill\COMMENT{propagate state $s$} 
    \STATE $V=\mbox{GramSchmidt}(V)$\hfill\COMMENT{re-orthogonalization}
    \STATE $\forall i\;:\;\lambda_i  = \lambda_i + \log_2 \bigl(||v_i||_2\bigr)$\hfill\COMMENT{accumulation of LE}
\ENDFOR
\RETURN $\bigl\{\frac{\lambda_i}{Tdt}\bigr\}_{i=1}^n$\hfill\COMMENT{Truncated Lyapunov Exponents}
\end{algorithmic}
\end{pseudocodeLE}

\section{Numerical Simulations}

\noindent The numerical experiments address the following questions:  1. Does SuPLE in Eq.~\eqref{eq:SuPLE} solve the task of stabilization from samples without auxiliary exploration? 2. Does just MaxLE already solve these tasks? 3. Is there a performance gap between traditional  rewards such as Quadratic and Sparse vs. the SuPLE reward?   

For this, we chose three dynamical systems of different complexity: `Single Pendulum':a simple dynamics, frequently used in evaluation of new methods; `Cart-Pole': a more complicated system, a prototype of the `Segway' robotic system; `Double Pendulum': a complicated chaotic dynamics with an extremely challenging exploration of the state space. It is a prototype for self-righting multi-linked robotic systems. 

All systems start at their stable equilibrium (e.g., the bottom position for the pendulum links) in both training and testing. Importantly, auxiliary exploration 
is used neither in training nor in testing. All the systems are trained with the same parameters except for the following reward functions:

\begin{itemize}
    \item `SuPLE': sum of all positive LE given by Eq.~\eqref{eq:SuPLE}
    \item `MaxLE': only the maximum of the positive LE
    \item `Quadratic': weighted error between the current state, $s$, and the target state, $s^g$: $r = - \sum_{i=1}^n \alpha_i (s^g_i - s_i)^2$. 
    \item `Sparse': $r=\begin{cases}
        1 &\text{if } \sum_{i=1}^n(s^g_i - s_i)^2<\epsilon\\
        0 &\text{otherwise}
    \end{cases}$.
\end{itemize}

In training we use the standard Soft-Actor Critic \cite{brockman2016openai} and the standard model parameters \cite{tiomkin2024intrinsic}. The dynamics models are provided in Appendix. All the results are averaged over ten models per system in training, and over five trajectories, in turn, per model in testing. Thus, mean and variance are calculated with 50 trajectories for each reward.

\subsection{Key Observations}
SuPLE and MaxLE do not require the specification of a target state, but instead
 guide the agent towards unstable equilibria, and then stabilize  the agent  there. 
The experiment designer does not need to know where these unstable equilibria occur; the agent will find them by itself. SuPLE allows for the derivation of a single policy for both self-righting (swinging up) and staying stable at the up-right position. 

In contrast, Quadratic and Sparse rewards do require an explicit specification of the target states, e.g., the upright position for self-righting robotic systems. In general, to specify unstable equilibria requires one to carry out a sensitivity analysis \cite{strogatz2018nonlinear}. Thus, in order to manually design effective reward, one needs a) to understand the properties of dynamics, and b) to explicitly calculate/specify the target states. In contrast, SuPLE reward does require neither `a)' nor `b)' in stabilization tasks.    

The experiments show that in the simple environments (Single Pendulum, Cart Pole) the task is solved by all the above-mentioned reward functions without auxiliary exploration. However, in the Double Pendulum,  exploration is challenging, and neither Sparse nor Error-based rewards succeed to solve the task, while SuPLE succeeds. Swinging up and stabilizing the Double Pendulum are prototypical primitives for self-righting multi-linked robots, making SuPLE a promising candidate for such in real robotic systems, where auxiliary exploration is unpractical.

Below, we show the SuPLE landscape  SuPLE and  the test error between the current state and the unstable equilibrium. The task is to learn, using samples, a controller for self-righting from the bottom with different rewards: externally-provided (Sparse and Quadratic) and intrinsically-calculated  (SuPLE and MaxLE).

\subsection{Experiments}

\subsubsection{\textbf{Simple Pendulum}}
This simple prototypical model for uprighting is the minmum benchmark test for new methods. It is a 2D system with its state at time $t$ given by $s(t)=[\theta, \dot{\theta}]$, $\theta$ being the pendulum angle with $\theta=\pi$ $rad$ at the top. The agent controls the systems by applying torque, $|a(t)|<1$ N/m, directly to the angular acceleration. 
Figure~\ref{fig:pendulum_sop_landscape} shows the SuPLElandscape,  providing the agent with a dense reward with the maximum at the top. For this simple system, all four rewards allow the agent to successfully learn a stabilizing policy without auxiliary exploration (resetting at arbitrary states) Figure~\ref{fig:pendulum_successrate}.
\begin{figure}[h!]
\centering
   \includegraphics[width=0.9\linewidth]{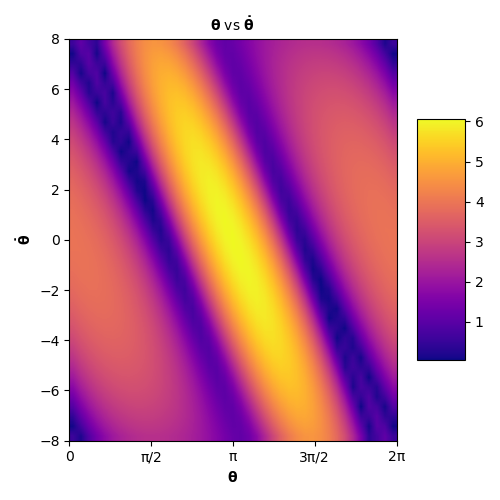}
   \caption{Landscape of SuPLE in Single Pendulum. x-axis: angle, y-axis: angular velocity}
   \label{fig:pendulum_sop_landscape} 
\end{figure}

\begin{figure*}[t!]
    \centering
    \begin{subfigure}[t]{0.32\textwidth}
        \centering
   \includegraphics[width=1\linewidth]{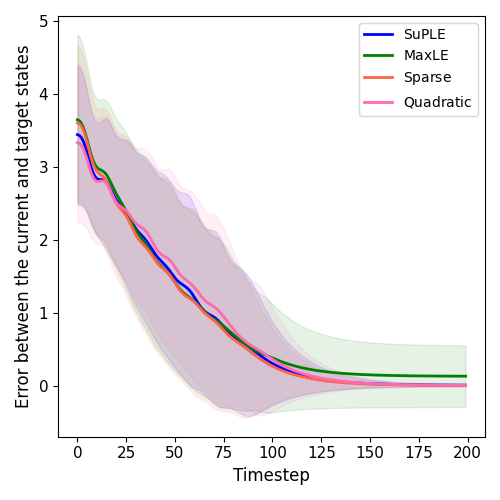}
   \caption{\textbf{Single Pendulum.}}
   \label{fig:pendulum_successrate} 
    \end{subfigure}%
    ~ 
    \begin{subfigure}[t]{0.32\textwidth}
        \centering
            \includegraphics[width=1\textwidth]{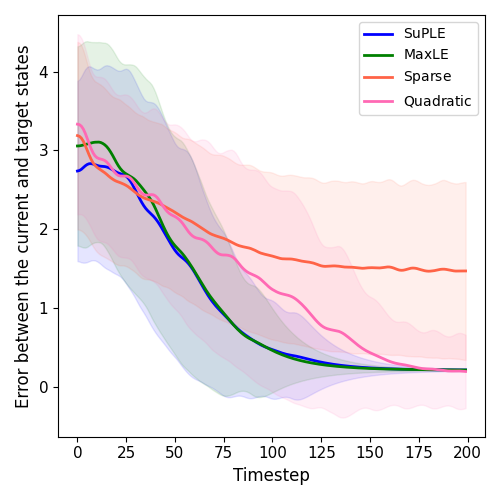}
            \caption{\textbf{Cart Pole.}}
            \label{fig:cartpole_successrate}
    \end{subfigure}
    ~ 
    \begin{subfigure}[t]{0.32\textwidth}
        \centering
\includegraphics[width=1\textwidth]{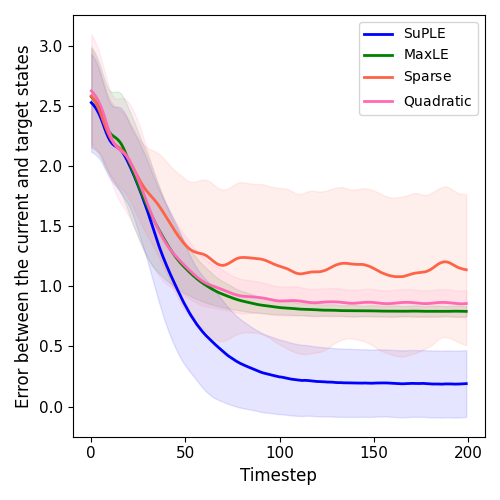}
\caption{\textbf{Double Pendulum.}}
\label{fig:dp_successrate}
    \end{subfigure}
    \caption{Test error to the upright position for different rewards \emph{without random position resetting} in training. In more complicated systems (left-to-right), Sparse and Error-Based (Quadratic) rewards fail, SuPLE succeeds in all systems.}
\end{figure*}

\begin{figure*}[t]
\begin{subfigure}[t]{1\textwidth}
\centering
\includegraphics[width=\textwidth]{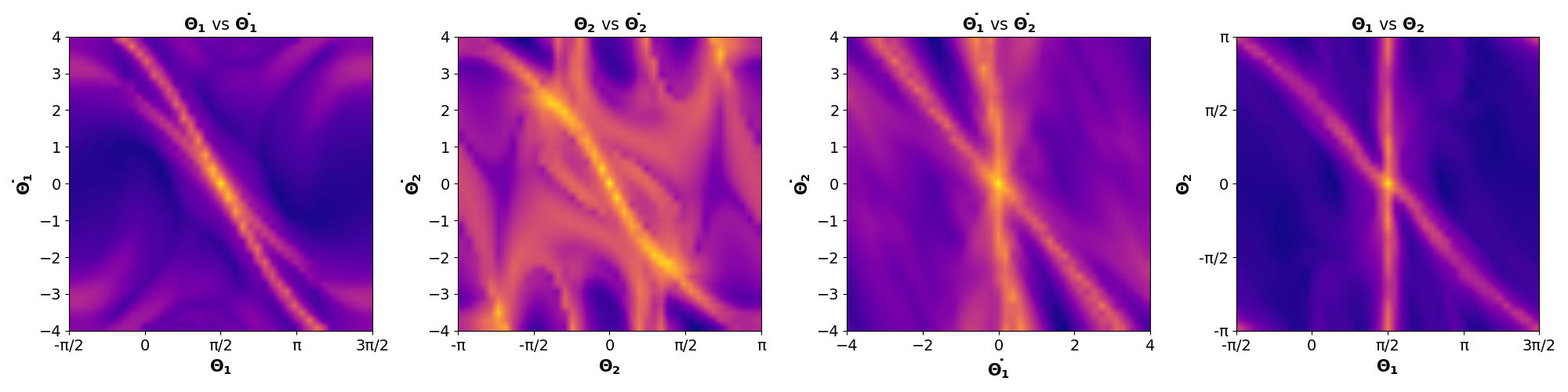}
\caption{\textbf{SuPLE Landscape.}}
\label{fig:dp_sop_landscape}
\end{subfigure}
\begin{subfigure}[t]{1\textwidth}
    \centering
\includegraphics[width=\textwidth]{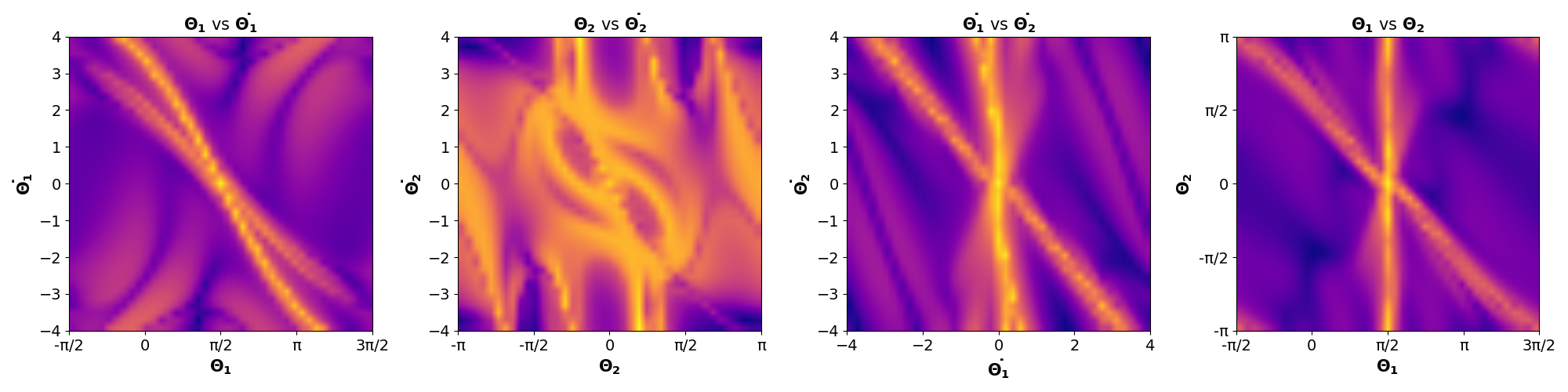}
\caption{\textbf{MaxLE Landscape.}}
\label{fig:dp_max_landscape}
\end{subfigure}
\caption{Pair-wise representation of the 4D landscape.  $\theta_1$, $\dot{\theta}_1$, $\theta_2$, and $\dot{\theta}_2$ are the angle of the first pole, its angular velocity, and analogously for the second pole.
When both poles are aligned at the top position $\theta_1 = \frac{\pi}{2}$ $rad$ and $\theta_2 = 0$ $rad$.%
}\label{fig:dp_landscapes}
\end{figure*}

\subsubsection{\textbf{Cart Pole}}
This system is more complicated due to the indirect control of the angular pole  by the agent's action (force) applied to the 
cart. This system is prototypical for `Segway'-style locomotion. Usually the standard RL methods \cite{brockman2016openai} train the swinging-up phase and the stabilization phase separately with two different rewards resulting in two different controllers, respectively. Here, to demonstrate the robustness and generality of SuPLE, we train the agent with the \emph{same} reward function without manually separating it into  two phases.

The SuPLE landscape for the angular coordinates is qualitatively similar to that of the `Single Pendulum'. The test error at Figure~\ref{fig:cartpole_successrate} shows that the non-informative sparse reward does not allow the agent to solve the task without  resetting `Cart Pole' at random arbitrary states in training, which is unpractical with a real system. Additionally, the agent achieves the upright position more slowly in comparison to the `Simple Pendulum', when trained with the externally provided Quadratic reward.

\subsubsection{\textbf{Double Pendulum}}
This is a complicated system with chaotic dynamics and a hard-to-explore state space. It is a  prototypical system for multi-linked systems with the task of self-righting. This experiment is to validate that SuPLE is able to solve this task in  realistic training conditions, when one resets to fixed rather than arbitrary states in multi-linked, e.g.\ humanoid, robots. As shown at Figure~\ref{fig:dp_successrate} amongst all rewards, only SuPLE is able to solve the task.  

The landscape of SuPLE and MaxLE in Figure~\ref{fig:dp_landscapes}, shows a marked difference between the two choices in the $\theta_2$ vs.\ $\dot\theta_2$ plane. Only SuPLE captures the volume expansion and has a markedly pronounced ridge along the upright-preferring states, whereas the MaxLE maxima are far more intricately distributed.

\begin{figure}[t!]
\centering
   \includegraphics[width=0.95\linewidth]{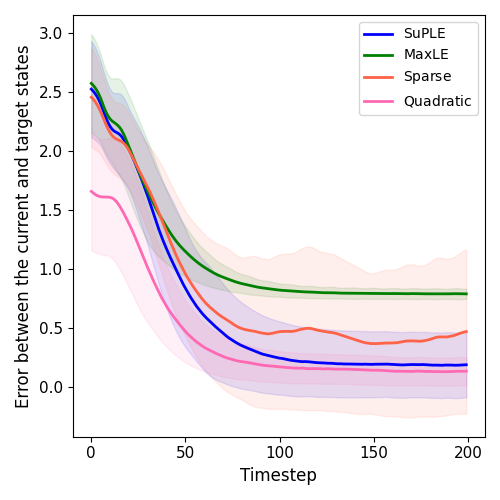}
   \caption{Double pendulum test error \textit{with random position resetting} in training.}
   \label{fig:dp_success_rate_random} 
\end{figure}

Finally, Figure~\ref{fig:dp_success_rate_random} shows that, with auxiliary exploration, the agent can solve the task with any of the rewards. 

}

\section{Discussion}

The truncated Lyapunov exponents as a reward signal
do not only stabilize the system near
its balancing points, but  produce  the full swing-up
process.
This method is superior to traditional RL, which
requires hand-crafted reward functions. In the latter, often much
effort needs to be invested into designing appropriate reward
functions which reach satisfactory results efficiently.
In problematic scenarios, either much domain knowledge needs to be
incorporated into the reward (the whole field of \emph{reward shaping}
\cite{ng_shaping}
); or else, one needs a suitably chosen
RL algorithm 
that propagates the reward signal swiftly throughout the policy domain.

All this is not required with the Lyapunov-based 
method. It uses inherent properties of the dynamics of the system to
``ride the crest of sensitivity'' towards the
points of maximum instability. It is plausible that this approach
 works in the neighbourhood of the
unstable points of the system: moving towards greater Lyapunov
exponents indicates a more unstable dynamics which, if bounded,
ultimately reaches a state of locally maximal instability.
Such states are often desired targets of a controller. A less
obvious property of the SupLE  method is, though, that the
``crests'' of sensitivity reach deeply into the state space. Thus, it is often possible to reach such a crest from
larger parts of the state space and then proceed to ride it all the
way towards the state of maximum sensitivity.
In other words, the basins of attraction for a controller based on
Lyapunov exponents are remarkably large. As for now, there
is no closed theory  for the size of these basins.

Creating essentially domain knowledge-free rewards through   SuPLE, however, 
comes at a price. In a traditional RL
formalism, the designer can make  specific states  desirable or undesirable,
independently of the properties of the dynamics, for instance,  tuning the rewards towards particular goal
areas that have  nothing to do with particularly unstable
 or otherwise distinguished points of the dynamics.

Lyapunov exponents can obviously not be used to induce such
arbitrarily imposed behaviours. However, in by far most typical use
cases, rewards are designed to achieve particular distinguished states
in the domain space, which the Lyapunov exponents seem particularly
suited to detect. Presently, this observation is made on a
purely empirical basis, but we highlight the links between SuPLE and the notion of \emph{empowerment}
\cite{salge2014empowerment, zhao2020efficient, qureshi2018adversarial}. The latter has been shown to be effective in a surprisingly wide range of control scenarios \cite{klyubin08:_keep_your_option_open}. 

We conjecture a further link. In a linear time-invariant control system, the Lyapunov exponents are linked to the the lower bound on the number of bits per time the controller needs to receive in order to stabilize the dynamics \cite{tatikonda2004control}. Thus, by maximizing the SuPLE reward, one might be find states where one can ``pry open" the channel through which a system could be stabilized.

{

In conclusion, we found that the Lyapunov Exponents provide ``natural'' rewards directly from system
properties. In particular,  no hand-designed reward-shaping is
necessary, as long as the desired tasks bear a relation to
``interesting'' (unstable) points of the dynamics.
Stable
points are not particularly interesting for control, since little to
no control is required to reach them. One other striking insight is
that the crests leading to the summit points of this Lyapunov-induced
reward landscape often reach far into the state space and thus provide
highly structured, informative and thus easy to learn value functions
throughout the state space

}

\section*{ACKNOWLEDGMENT}
PN was supported in part by the RSCA scholarship at SJSU; DP was supported in part by Pazy grant (195-2020); ST was supported in part by NSF Award (2246221) and Pazy grant (195-2020).



\bibliographystyle{IEEEtran}
\bibliography{ref}

\addtolength{\textheight}{-12cm}   



\end{document}